\documentclass[journal,twoside,web]{ieeecolor}
\usepackage{generic}
\usepackage{cite}
\usepackage{amsmath,amssymb,amsfonts}
\usepackage{float}
\usepackage{algorithmic}
\usepackage{graphicx}
\usepackage{algorithm,algorithmic}
\usepackage{hyperref}
\hypersetup{hidelinks=true}
\usepackage{textcomp}
\usepackage{booktabs}
\usepackage{subcaption}

\def\BibTeX{{\rm B\kern-.05em{\sc i\kern-.025em b}\kern-.08em
		T\kern-.1667em\lower.7ex\hbox{E}\kern-.125emX}}
\usepackage{amsmath}
\usepackage{amssymb}

\usepackage{float}

\usepackage{caption}

\usepackage{multirow}

\usepackage{array}
\newcolumntype{C}{ >{\centering\arraybackslash} m{1.0cm} }
\newcolumntype{Q}{ >{\centering\arraybackslash} m{3 cmcm} }
\newcolumntype{M}[1]{>{\centering\arraybackslash}m{#1}}
\newcolumntype{P}[1]{>{\centering\arraybackslash}p{#1}}

\begin{document}
	
	\title{ReACT-Drug: Reaction-Template Guided Reinforcement Learning for \textit{de novo}\\ Drug Design}
	\author{R Yadunandan  and Nimisha Ghosh \IEEEmembership{Member, IEEE}
		\thanks{R Yadunandan and Nimisha Ghosh are with the Department of Computer Science and Engineering, Shiv Nadar University Chennai, Chennai, Tamil Nadu, India (e-mail: nimishaghosh@snuchennai.edu.in). }% <-this % stops a space
		\thanks{Manuscript received April 19, 2005; revised August 26, 2015.}}
	\markboth{Journal of \LaTeX\ Class Files,~Vol.~14, No.~8, August~2015}%
	{Ghosh \MakeLowercase{\textit{et al.}}: Bare Demo of IEEEtran.cls for IEEE Journals}
	
	\maketitle

\begin{abstract}
\textit{De novo} drug design is a crucial component of modern drug development, yet navigating the vast chemical space to find synthetically accessible, high-affinity candidates remains a significant challenge. Reinforcement Learning (RL) enhances this process by enabling multi-objective optimization and exploration of novel chemical space -  capabilities that traditional supervised learning methods lack. In this work, we introduce \textbf{ReACT-Drug}, a fully integrated, target-agnostic molecular design framework based on Reinforcement Learning. Unlike models requiring target-specific fine-tuning, ReACT-Drug utilizes a generalist approach by leveraging ESM-2 protein embeddings to identify similar proteins for a given target from a knowledge base such as Protein Data Base (PDB). Thereafter, the known drug ligands corresponding to such proteins are decomposed to initialize a fragment-based search space, biasing the agent towards biologically relevant subspaces. For each such fragment, the pipeline employs a Proximal Policy Optimization (PPO) agent guiding a ChemBERTa-encoded molecule  through a dynamic action space of chemically valid, reaction-template-based transformations. This results in  the generation of \textit{de novo} drug candidates with competitive binding affinities and high synthetic accessibility, while ensuring 100\% chemical validity and novelty as per MOSES benchmarking. This architecture highlights the potential of integrating structural biology, deep representation learning, and chemical synthesis rules to automate and accelerate rational drug design. The dataset and code are available at https://github.com/YadunandanRaman/ReACT-Drug/.
\end{abstract}


\section{Introduction}
\label{sec:introduction}
Development of a new drug where novel drug molecules are created is a time consuming and a very costly process. For such drug design, the number of feasible synthetic molecules is of the order of $10^{60}-10^{100}$, out of which the suitable compound needs to be identified that should satisfy properties like bioactivity, drug metabolism and pharmacokinetic profile etc.~\cite{Olivecrona2017}. This in itself is a very gigantic task and to mitigate such an issue, computational approaches can be very useful for \textit{de novo} drug design which can help in significantly reducing the expenses and time of drug development.

In the recent past, many deep learning algorithms have contributed significantly in molecule generation. In this regard, generative models have played a very important role wherein the models capture the inherent molecular distribution of the training dataset and then use this learned distribution to generate new molecules. AutoEncoder- (AE)~\cite{Blaschke2018, Polykovskiy2018} and Variational AE-based~\cite{Joo2020, Bhadwal2024, Gomez2018, Kang2019, yoshikai2024, Ochiai2023} models transform training molecules to a latent space using an encoder, generate new molecules in that latent space and then reconstructs the molecules using a decoder. Also, Generative Adversarial Network (GAN)~\cite{Macedo2024, Wang2025} based models have also shown quite good results in molecule generation. Furthermore, Transformer-based generators~\cite{Bagal2022,Liu2024}, Diffusion-based generators~\cite{Zhang2025, Qin2025} and Recurrent Neural Network~\cite{ZHU2025,Kotsias2020}-based generators have also shown immense potential and promising results. 

In the recent past, works like \cite{Zhou2019} have utilized Reinforcement Learning (RL) to generate drug-like molecules that are similar to the original molecule while \cite{Fang2023} generate drug molecules from scratch. RL has shown promising results for \textit{de novo} molecular generation where the objective is learning a policy than can sample sequences of tokens to generate simplified molecular-input line-entry system (SMILE) strings. RL methods with sequence-based approaches are also quite popular for controlling SMILES generation. These approaches employ the benefits of large language models that are pretrained on a huge number of chemical datasets, thereby capturing the chemical language quite well and providing encouraging results~\cite{Mazuz2023,Olivecrona2017}. In~\cite{Haddad2025}, the authors have proposed MOLRL ((Molecule Optimization with Latent Reinforcement Learning) that uses an RL paradigm called Proximal Policy Optimization (PPO)~\cite{schulman2017} to optimize molecules in the latent space of a pretrained generative model. This framework is agnostic to the architecture of the generative model although the characteristics of the latent space have an important effect on the optimization method. In~\cite{Svensson2024}, the authors explore many policy optimization algorithms including PPO for \textit{de novo} drug design in order to generate different types of molecules with high scores. Gupta et al.~\cite{Gupta2025} have proposed Policy-guided Unbiased REpresentations (PURE) which employs semi-supervised and agnostic training design along with template-based molecular simulations for Structure-Constrained Molecular Generation (SCMG) tasks. SCMG is a method that uses existing molecules to guide the creation of new ones with similar structures but improved properties. 

Motivated by the literature, in this work, we propose the design of a \textit{de novo} molecular drug discovery system based on an RL pipeline. It integrates ESM-2~\cite{Lin2023} for protein embedding and ChemBERTa~\cite{chithrananda2020} for molecular representation which is a fragment-based approach for initial state generation, a novel reaction template library for molecular transformations, and an \textit{in silico} docking oracle (AutoDock Vina) for reward calculation. The entire workflow is managed by a PPO agent designed to navigate a dynamic action space while optimizing molecules toward multiple objectives.

Our contribution can be summarized as follows:
\begin{itemize}
    \item We propose a novel target-agnostic, policy-based reinforcement learning technique viz. ReACT-Drug for \textit{de novo} drug design for a target protein by utilizing a fragment-based state generation.
    \item ReACT-Drug eliminates the need for extensive supervised pre-training on specific target-drug pairs, relying instead on pre-trained generalist encoders and episodic reinforcement learning.
    \item The experimental results demonstrate that the proposed framework shows superior or competitive performance as compared to the existing models.
\end{itemize}
\section{Materials and Methods}
\begin{figure*}  
    \centering
    \includegraphics[width=7.0in]{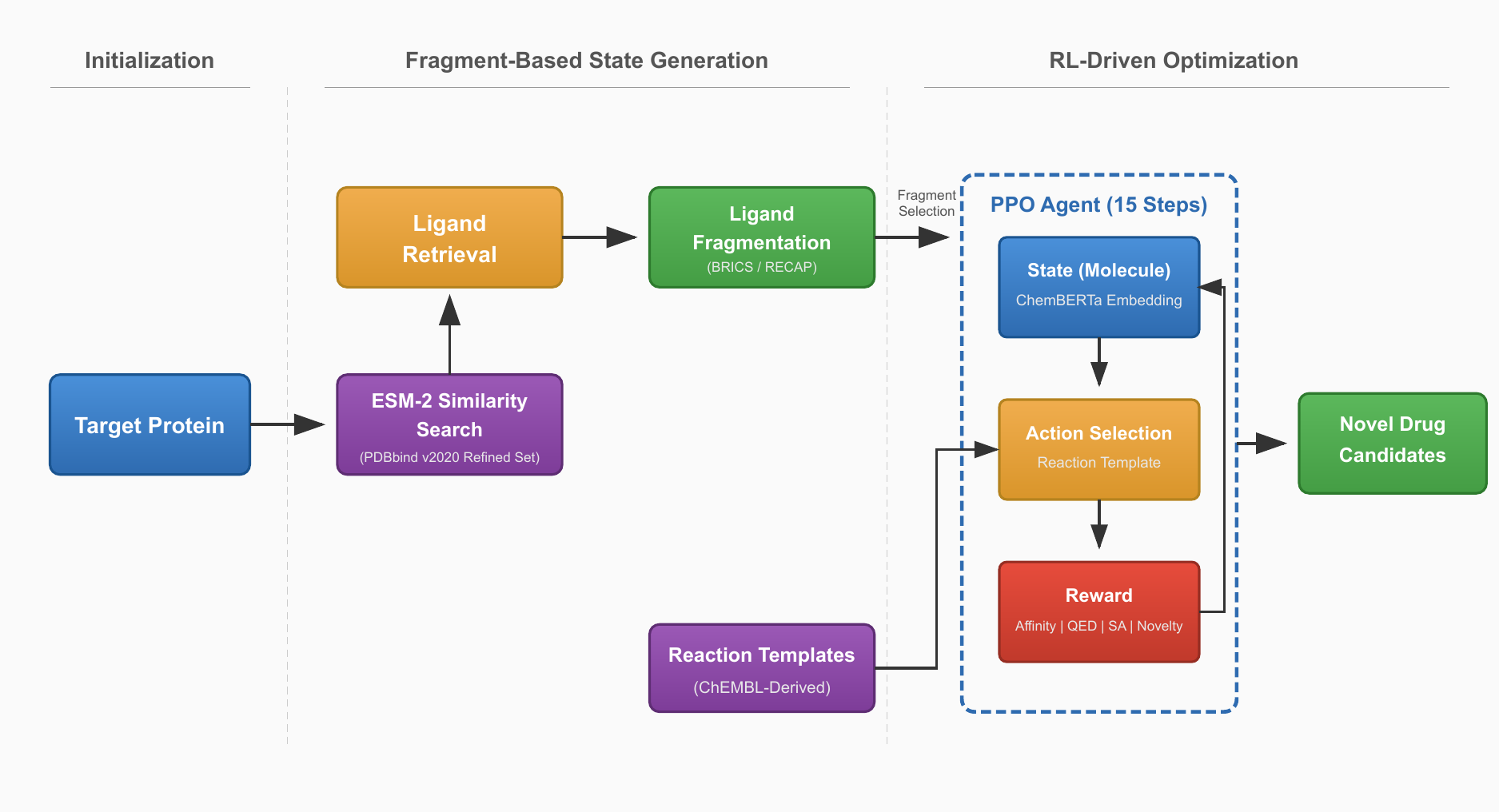}
    \caption{\textit{De novo} molecular drug discovery pipeline workflow. The system initializes with a target protein, identifies similar proteins using ESM-2 embeddings, fragments their ligands, and uses a PPO agent to iteratively optimize molecules over 15-step episodes. The agent selects chemical transformations from a drug-relevant template library, with rewards based on binding affinity (AutoDock Vina), drug-likeness (QED), synthetic accessibility, and novelty.}
    \label{fig:workflow}
\end{figure*}

In this work, the discovery process begins by defining a target protein (PDB structure, sequence, and binding site) and subsequently the system identifies known drug ligands for functionally similar proteins using ESM-2 embeddings. These ligands are decomposed into a pool of chemical fragments. An RL episode starts by randomly selecting one fragment, which the PPO agent iteratively ``grows'' by applying chemically valid transformations from a pre-compiled library. At each step, the agent's action (the choice of transformation) is guided by a policy network. The resulting molecule is evaluated by a multi-objective reward function, dominated by its docked binding affinity to the target, which is calculated using AutoDock Vina. The agent learns to select transformations that maximize this multi-objective reward, ultimately generating novel, high-affinity, and drug-like molecules. All computational experiments, including the training of the PPO agent and the docking simulations, are conducted using an NVIDIA L4 GPU. Figure~\ref{fig:workflow} illustrates the overall workflow of the system, from protein target initialization to RL-driven molecular optimization.

\subsection{Data and Model Preparation}

\subsubsection{Protein-Ligand Knowledge Base}

The system leverages the PDBbind v2020 refined dataset as its primary source of protein-ligand interaction data. This dataset is processed to extract protein sequences, canonical ligand SMILES strings, and experimental binding affinity values. This processed data serves as a ``knowledge base'' for identifying relevant starting molecules.

\subsubsection{Reaction Template Library Generation}

A critical component of the system is its action space, which consists of a large library of chemically valid molecular transformations. This library is not based on general synthesis datasets  but is instead derived from known drug-like molecules.

The reaction template library is generated using the following workflow:

\begin{enumerate}
\item \textbf{Data Acquisition:} A large corpus of bioactive molecules is downloaded from the ChEMBL database and converted to SMILES format. The molecules are preprocessed using the mmpdb (Matched Molecular Pair Database)~\cite{Dalke2018} tool through two stages: fragmentation to identify substructures, and indexing to create a queryable database of molecular transformations.

\item \textbf{MMP Analysis:} The mmpdb database is queried to extract all transformation rules between fragment pairs. Each rule represents a structural transformation (for example, substituting a hydrogen for a fluorine, \texttt{[*:1]H >> [*:1]F}), along with its frequency of occurrence across the ChEMBL dataset. Rules are aggregated by their SMARTS (SMiles ARbitrary Target Specification) patterns and ranked by frequency.

\item \textbf{Filtering:} The resulting transformations are heavily filtered to retain only those considered ``drug-relevant'' using the following criteria:
\begin{itemize}
    \item Maximum of 10 heavy atoms in either variable fragment
    \item Variable size must not exceed 50\% of the parent molecule size
    \item Molecular core must contain at least six heavy atoms
    \item Minimum frequency of one occurrence in the database
\end{itemize}
These constraints ensure transformations are conservative modifications suitable for systematic exploration of chemical space around drug-like molecules.

\item \textbf{Template Creation:} The filtered transformations are converted into reaction SMARTS templates (for example, \texttt{[*:1][H]>>[*:1][F]}). Each template preserves both the structural pattern and frequency metadata, enabling prioritization of commonly observed medicinal chemistry transformations.
\end{enumerate}

This process yields a final library of thousands of high-confidence, drug-relevant reaction templates, which function as the agent's discrete action space.

\subsubsection{Protein and Molecule Encoders}

Two pre-trained transformer models are used for feature extraction:

\begin{itemize}
\item \textbf{Protein Encoder:} Protein embeddings are obtained using ESM-2~\cite{Lin2023} model with 650 M parameters to generate fixed-dimensional embeddings (1280-dim) from protein amino acid sequences.

\item \textbf{Molecule Encoder:} Molecular embeddings are derived from ChemBERTa \cite{chithrananda2020} model to generate embeddings (768-dim) from molecular SMILES strings. These embeddings serve as the state representation for the RL agent.
\end{itemize}

\subsection{Target-Specific Environment Initialization}

\subsubsection{Target Definition and In Silico Oracle}

The user provides the target's PDB structure file and a custom AutoDock Vina interface module initializes an \textit{in silico} oracle by preparing the protein structure. This structure is converted into AutoDock-compatible format (.pdbqt) using Open Babel
%~\cite{Boyle2011}. 

% Any molecule generated by the agent can then be passed to this module, which prepares the ligand, runs the vina executable, and parses the output to return the predicted binding affinity (kcal/mol).

\subsubsection{Fragment-Based Starting State Generation}

To bias the discovery process of a \textit{de novo} drug for a relevant target, the system generates a unique pool of starting molecules for each such target.

\begin{enumerate}
\item \textbf{Protein Similarity Search:} The target's protein sequence is encoded using ESM-2. This embedding is used to perform a cosine similarity search against all protein sequences in the PDBbind knowledge base, retrieving the top-$k$ ($\mathcal{N}_k$) most similar proteins. This can be represented mathematically as:
\begin{equation}
    \mathcal{N}_k = \underset{i}{\operatorname{arg\,top\text{-}}k} \frac{f_{ESM}(S_t).f_{ESM}(S_i)}{\left\| f_{\text{ESM}}(S_t) \right\|_2
\left\| f_{\text{ESM}}(S_i) \right\|_2}
\end{equation}
where, $f_{ESM}(S_t)$ and $f_{ESM}(S_i )$ are the target protein sequence and the protein sequence in PDBbind respectively while $f_{ESM}(.)$ denotes the ESM-2 embedding function.

\item \textbf{Ligand Retrieval:} The known drug ligands ($\mathcal{L}$) for these similar proteins are collected.
\begin{equation}
    \mathcal{L} =
\bigcup_{i \in \mathcal{N}_k} \mathcal{L}_i
\end{equation}

\item \textbf{Fragmentation:} These parent ligands are decomposed into smaller chemical fragments ($\mathcal{F}$) using RDKit's BRICS (Breaking of Retrosynthetically Interesting Chemical Substructures)~\cite{Degen2008} and RECAP (Retrosynthetic Combinatorial Analysis Procedure)~\cite{Lewell1998} algorithms.
\begin{equation}
    \mathcal{F}
=
\bigcup_{\ell\in\mathcal{L}}
\left(
\mathrm{BRICS}(\ell)
\cup
\mathrm{RECAP}(\ell)
\right)
\end{equation}

\item \textbf{Pool Creation:} This collection of fragments (for example, \texttt{c1ccccc1}, \texttt{CCO} etc.) forms the starting pool of molecules. An RL episode begins by randomly selecting one fragment from this pool,
\end{enumerate}

\begin{equation}
    M_0 = f_0, \qquad 
f_0 \sim p(f),\; f \in \mathcal{F}.
\end{equation}
This approach allows the agent to ``grow'' a molecule from a relevant seed rather than starting from scratch. Fragment-based initialization constrains exploration to biologically relevant subspaces, accelerating convergence.

\subsection{Reinforcement Learning for Molecular Generation}

The core of the system is a PPO agent trained to iteratively modify the selected fragment over a fixed-length episode which runs for 15 steps to maximize a multi-objective reward. It is important to note here that each episode starts with the selection of a fragment and terminating with the calculation of the reward function at the end of the 15th step.

\subsubsection{Markov Decision Process (MDP) Formulation}

\begin{itemize}
\item \textbf{State ($s_t$):} The 768-dim ChemBERTa embedding of the molecule at step $t$.
\begin{equation}
    s_t = \mathbf{h}(M_t) \in \mathbb{R}^{768}
    \label{eq:5}
\end{equation}
where $\mathbf{h}(\cdot)$ denotes the ChemBERTa encoder and $M_t$ is the
molecule at step $t$.

\item \textbf{Action ($a_t$):} The index of a reaction template selected from a dynamically generated list of applicable templates.
\begin{equation}
    \mathcal{A}(M_t) = \{ \tau_1, \tau_2, \ldots, \tau_{K_t} \} 
\end{equation}
is the dynamically generated set of reaction templates applicable to $M_t$.
The action selected at time $t$ is:
\begin{equation}
    a_t = \tau_{i_t}, \qquad i_t \in \{1, \ldots, K_t\}
\end{equation}
\item \textbf{Transition ($s_{t+1}$):} The application of the chosen template to the current molecule $s_t$, producing a new molecule $s_{t+1}$. 
\begin{equation}
    M_{t+1} = T(M_t, a_t)
\end{equation}
where \(T(\cdot,\cdot)\) applies the selected reaction template.

The next state is:
\begin{equation}
    s_{t+1} = \mathbf{h}(M_{t+1})
\end{equation}
\item \textbf{Reward ($R_t$):} The final reward is calculated by a weighted sum of multiple objectives where the agent's objective is to find a policy $\pi_\theta$ that maximizes the expected cumulative discounted reward:
\begin{equation}
J(\theta) = \mathbb{E}_{\pi_\theta} \left[\sum_{t=0}^{T} \gamma^t R_t \right]
\end{equation}
where, $\gamma \in (0,1]$ is the discount factor, $T$ is the number of horizon and $R_t$ is the scalarized reward at step $t$:
\begin{equation}
\begin{aligned}
R_t &= 
w_1 R_{\text{affinity}}(M_t) 
+ w_2 R_{\text{QED}}(M_t) \\
&\quad - w_3 R_{\text{SA}}(M_t) 
+ w_4 R_{\text{novelty}}(M_t)
\end{aligned}
\end{equation}
\end{itemize}
The objectives include binding affinity (from AutoDock Vina), drug-likeness (QED), synthetic accessibility (SA), and novelty. The values of the hyperparameter $w_1$, $w_2$, $w_3$ and $w_4$ are chosen after extensive experiments and are taken as 1, 0.1, 0.1 and 0.35 respectively.

\subsubsection{Dynamic Action Space}

At each step $t$, the action space is dynamic. The system takes the current molecule $M_t$ and iterates through the entire library of ChEMBL-derived reaction templates to determine which templates are chemically applicable to the current molecule. This creates a list of valid next molecules ($\mathcal{A}(M_t)$) , which form the action space for that specific step.

\subsubsection{PPO Agent and Action Selection}

We employ a Proximal Policy Optimization (PPO) agent with a custom actor-critic network. We chose PPO for its stability in dynamically sized action spaces and its sample efficiency relative to other on-policy methods. To handle the dynamic action space, the policy network is designed as follows:

\begin{itemize}
\item \textbf{Network Input:} The agent's network receives the current state (ChemBERTa embedding of molecule $M_t$, denoted by $s_t$ in Equation~\ref{eq:5}).

\item \textbf{Network Output:} The network outputs two heads: a value (predicting the state's quality ($v_t$)) and a policy query vector ($q_t$). Thus, based on the molecular embedding, the PPO policy network produces:
$$(v_t,q_t) = f_\theta(s_t)$$

\item \textbf{Action Selection:}
\begin{enumerate}
\item[a.] The SMILES for all valid next molecules (the dynamic action space) are encoded using ChemBERTa to get their respective embeddings; $e_i = h(M_{t+1}^i), i = 1,\dots, K_t$ and $K_t= |\mathcal{A}(M_t)|$. This creates a set of action embeddings denoted by $E_t = \{e_1,\dots,e_{K_t}\}$.

\item[b.] The agent's policy query vector ($q_t$) is compared (via dot product) to the embeddings  ($e_i$) of all possible next molecules. This operation can be represented by $z_i = q_te_i, i =1,\dots, K_t$.

\item[c.] This comparison produces logits, which are converted into a probability distribution given by:
\begin{equation}
    \pi_\theta(a_t = M_{t+1}^{(i)}|s_t) = \frac{exp(z_i)}{\sum_{j=1}^{K_t} exp(z_j)}
\end{equation}
\item[d.] The agent samples an action (a specific next molecule) from this distribution.
\end{enumerate}
\end{itemize}

This architecture allows the agent to generalize its policy $\pi_\theta$ across different states and varying action spaces, effectively learning ``which kind of transformation to apply'' rather than learning a fixed-size policy. The agent's experiences are stored in a buffer, and the PPO algorithm is used to update the network weights to maximize the expected cumulative reward.

\subsection{Discovery and Analysis}

The system is run for a specified number of episodes. Any molecule generated during this process that exceeds a pre-defined threshold for binding affinity and drug-likeness is collected as a ``discovery.'' The final output is a ranked list of these novel molecules, providing promising, synthetically accessible, and high-affinity candidates for the specified protein target.

\subsection{Illustrative Example}
\label{sec:example_workflow}

\begin{figure}
    \centering
    \includegraphics[width=0.85\columnwidth]{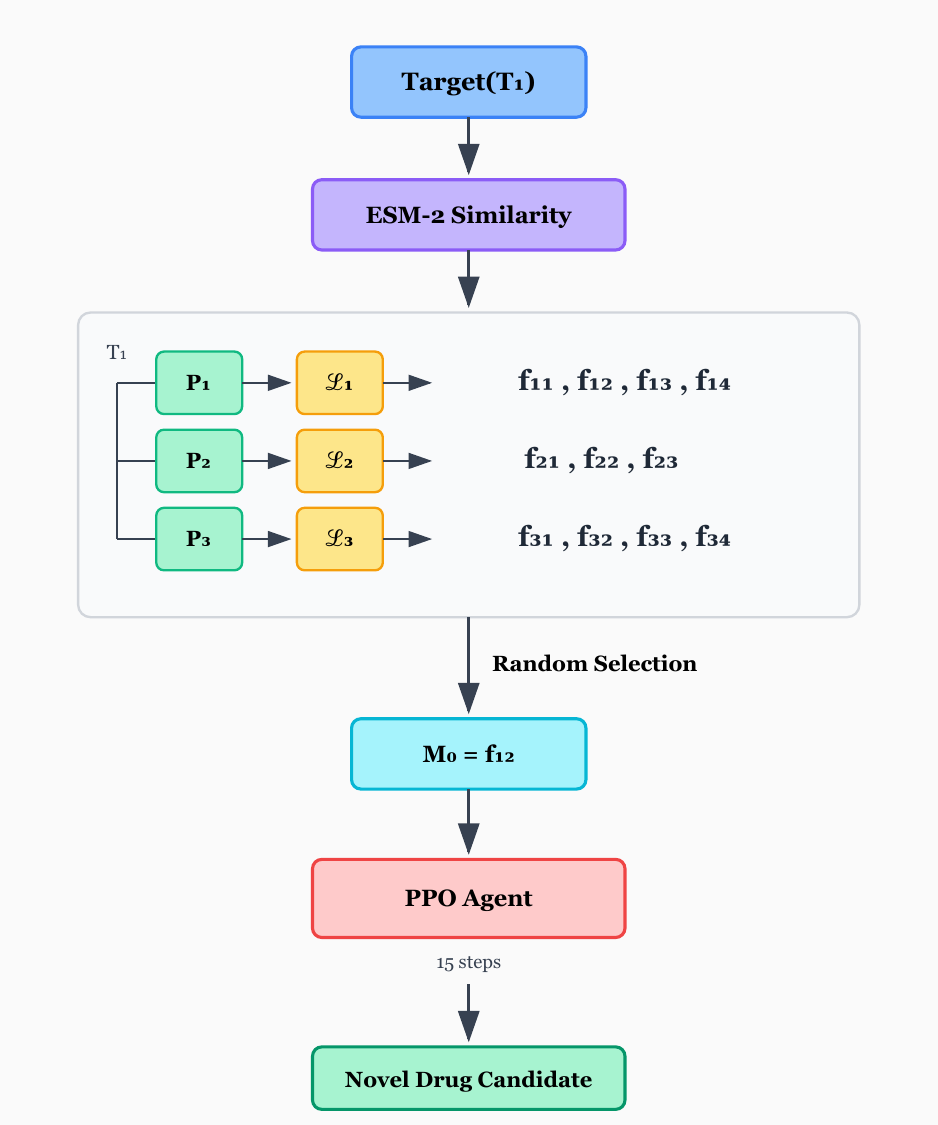}
    \caption{Illustrative example of the ReACT-Drug pipeline.}
    \label{fig:example_workflow}
\end{figure}

Figure~\ref{fig:example_workflow} illustrates the end-to-end ReACT-Drug pipeline. Given a target protein $T_1$, ESM-2 embeddings identify similar proteins $\{P_1, P_2, P_3\}$ from PDBbind. Their ligands $\{\mathcal{L}_1, \mathcal{L}_2, \mathcal{L}_3\}$ are decomposed into fragments $\{f_{ij}\}$ (denoted by $f_{11}, f_{12}$ etc.) via BRICS/RECAP . A randomly selected fragment $f_{11}$ initializes $M_0$, which the PPO agent iteratively transforms over 15 steps to yield a novel drug candidate.

\section{Results}
\label{sec:results}

A key advantage of the proposed ReACT-Drug framework is its target-agnostic design—the model requires no target-specific training or fine-tuning. To demonstrate this generalization capability, we evaluated our framework across six therapeutically diverse protein targets spanning different receptor families: serotonin receptors 5-HT1B (4IAQ), 5-HT2B (4NC3), muscarinic M2 receptor (3UON), kinase AKT1 (4GV1), dopamine D2 receptor (6CM4), and kappa opioid receptor (4DJH). This diverse selection tests the framework's ability to adapt to varying binding pocket geometries and pharmacological profiles without retraining.

\subsection{Properties of Generated Molecules}

Table~\ref{tab:mol_properties} summarizes the physicochemical properties of molecules generated by ReACT-Drug framework across all six protein targets. All generated molecules have achieved 100\% chemical validity and 100\% novelty, confirming that the reaction template-guided approach successfully produces chemically sound structures not present in the training data.

\begin{table}[htbp]
  \centering
  \caption{Physicochemical properties of generated molecules across six protein targets. Novelty is calculated with respect to the MOSES dataset~\cite{Polykovskiy2020}. \textbf{MW}: Molecular Weight (Da), \textbf{HBD}: Hydrogen Bond Donors, \textbf{HBA}: Hydrogen Bond Acceptors.}
  \label{tab:mol_properties}
  \begin{tabular}{@{}lccccc@{}}
    \toprule
    \textbf{Target} & \textbf{Valid} & \textbf{Novelty} & \textbf{MW} & \textbf{HBD} & \textbf{HBA} \\
    \midrule
    5-HT1B (4IAQ) & 1.0 & 1.0 & 501.93 & 1.67 & 8.33 \\
    5-HT2B (4NC3) & 1.0 & 1.0 & 489.93 & 1.33 & 4.67 \\
    M2 (3UON)     & 1.0 & 1.0 & 482.16 & 0.67 & 3.67 \\
    AKT1 (4GV1)   & 1.0 & 1.0 & 483.40 & 1.00 & 4.57 \\
    DRD2 (6CM4)   & 1.0 & 1.0 & 463.59 & 3.00 & 5.00 \\
    KOR (4DJH)    & 1.0 & 1.0 & 503.46 & 2.00 & 5.00 \\
    \midrule
    \textbf{Average} & \textbf{1.0} & \textbf{1.0} & \textbf{487.41} & \textbf{1.61} & \textbf{5.21} \\
    \bottomrule
  \end{tabular}
\end{table}

The generated molecules exhibit molecular weights in the range of 463-503 Da, which falls within the acceptable range for oral bioavailability according to Lipinski's rule of five (MW $<$ 500 Da)~\cite{LIPINSKI2004}. The hydrogen bond donor (HBD) counts range from 0.67 to 3.00, and hydrogen bond acceptor (HBA) counts range from 3.67 to 8.33, with most targets showing compliance with drug-likeness criteria (HBD $\leq$ 5 and HBA $\leq$ 10 (Lipinski’s rule of five)). We have also reported the physicochemical property profiles of the generated molecules in Figure~\ref{fig:radar_chart} which illustrates the distinct chemical ``fingerprints'' required for each binding pocket.

\begin{figure}
  \centering
  \includegraphics[width=0.95\columnwidth]{property_radar.pdf}
  \caption{\textbf{Physicochemical property profiles of generated molecules across six diverse targets.}  While Molecular Weight (\textbf{MW}) remains consistent across targets (clustered at $\approx 0.8$ normalized units), other properties like Hydrogen Bond Acceptors (\textbf{HBA}) and Donors (\textbf{HBD}) vary significantly (for example, high HBA demand for 5-HT1B vs. high HBD for DRD2), demonstrating the model's capacity to adapt to target-specific constraints. Values are min-max normalized for visual comparison.}
  \label{fig:radar_chart}
\end{figure}

\subsection{Drug-likeness and Synthetic Accessibility}
\begin{figure*}[t] % Use figure* for full-width; [t] usually works best for placement
    \centering
    \includegraphics[width=0.85\textwidth]{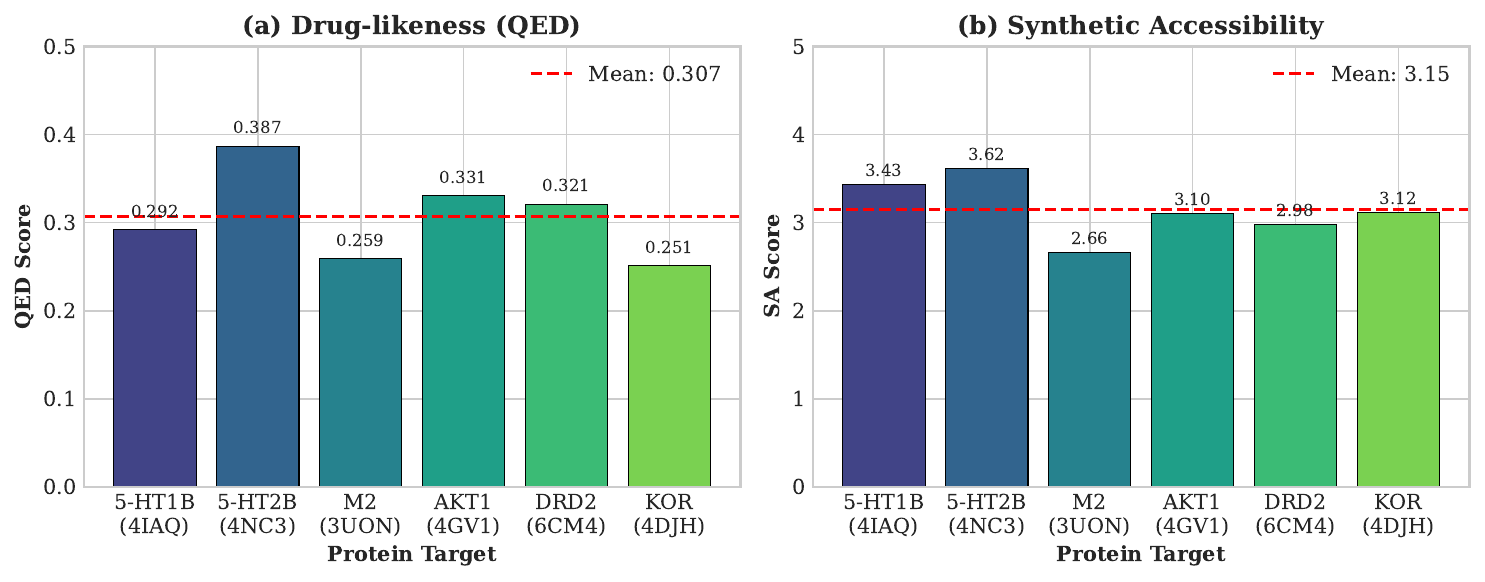} 
    \caption{Comparison of (a) QED scores and (b) Synthetic Accessibility scores across six protein targets. Higher QED indicates better drug-likeness, while lower SA scores indicate easier synthesis.}
    \label{fig:qed_sa}
\end{figure*}

Table~\ref{tab:druglikeness} and Figure~\ref{fig:qed_sa} show the distribution of QED scores and synthetic accessibility across all targets. 
Quantitative Estimate of Drug-likeness (QED) quantifies how similar a molecule is to known successful oral drugs.
The QED scores range from 0.251 (KOR) to 0.387 (5-HT2B), with an average of 0.307. While these values are moderate, they reflect the inherent trade-off between optimizing for high binding affinity and maintaining optimal drug-like properties—a well-documented challenge in multi-objective molecular optimization.

\begin{table}
\centering
\caption{Drug-likeness metrics of generated molecules across six protein targets.}
\label{tab:druglikeness}
\begin{tabular}{@{}lcc@{}}
\toprule
\textbf{Target} & \textbf{QED} & \textbf{SA Score} \\
\midrule
5-HT1B (4IAQ) & 0.292 & 3.43 \\
5-HT2B (4NC3) & 0.387 & 3.62 \\
M2 (3UON) & 0.259 & 2.66 \\
AKT1 (4GV1) & 0.331 & 3.10 \\
DRD2 (6CM4) & 0.321 & 2.98 \\
KOR (4DJH) & 0.251 & 3.12 \\
\midrule
\textbf{Average} & \textbf{0.307} & \textbf{3.15} \\
\bottomrule
\end{tabular}
\end{table}

Synthetic accessibility (SA) scores which tells us how difficult it is to synthesize a molecule in practice, range from 2.66 to 3.62, with an average of 3.15. These scores fall within the moderately synthesizable range (SA scores of 1-10, where lower is easier), suggesting that the generated molecules, while complex enough to achieve high binding affinity, remain within practical synthetic reach. The M2 receptor (3UON) target yielded molecules with the lowest SA score (2.66), indicating relatively straightforward synthetic routes.

\subsection{Target-Agnostic Performance Analysis}

\begin{figure}
    \centering
    \includegraphics[width=\columnwidth]{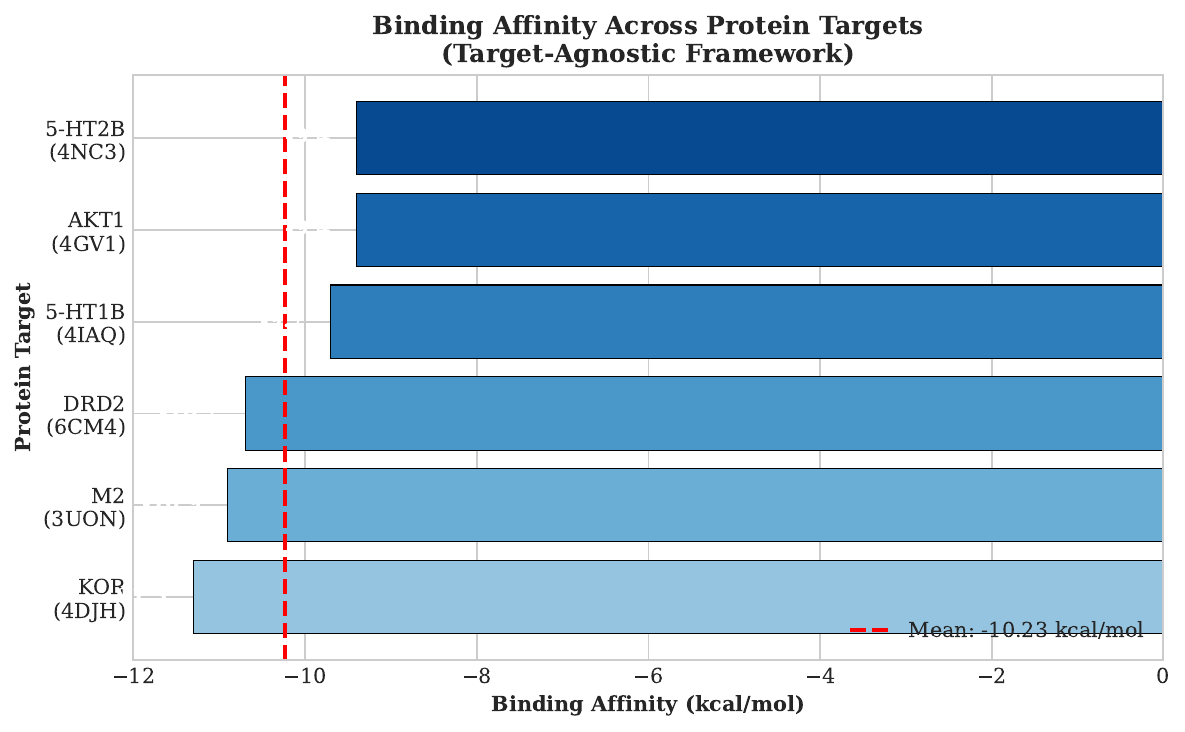}
    \caption{Binding affinity scores (kcal/mol) achieved by our target-agnostic framework across six diverse protein targets. More negative values indicate stronger predicted binding.}
    \label{fig:affinity_bar}
\end{figure}

Figure~\ref{fig:affinity_bar} summarizes the binding affinity performance across all evaluated targets. The consistent performance across diverse receptor families—GPCRs (5-HT1B, 5-HT2B, M2, DRD2, KOR) and kinases (AKT1)—demonstrates the robust generalization capability of our target-agnostic approach. This is a significant advantage over existing methods that often require extensive retraining or fine-tuning for each new target.

The framework's ability to leverage ESM-2 protein embeddings for identifying functionally similar proteins, combined with the ChEMBL-derived reaction template library, enables rapid deployment to novel therapeutic targets without the computational overhead of target-specific model training. This characteristic makes ReACT-Drug particularly suitable for early-stage drug discovery campaigns where rapid exploration of multiple target hypotheses is essential.

\subsection{Binding Affinity Comparison}

\begin{figure*}
    \centering
    \begin{subfigure}[b]{0.32\textwidth}
        \centering
        \includegraphics[width=\textwidth]{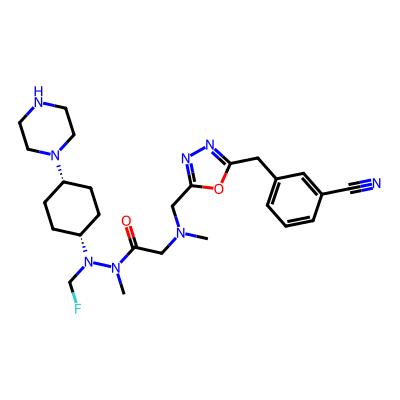}
        \caption{5-HT1B (4IAQ, -9.7 kcal/mol)}
    \end{subfigure}
    \hfill
    \begin{subfigure}[b]{0.32\textwidth}
        \centering
        \includegraphics[width=\textwidth]{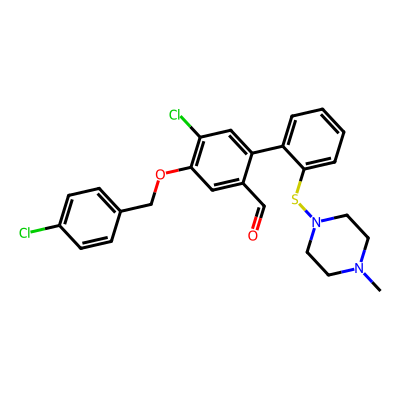}
        \caption{5-HT2B (4NC3, -9.4 kcal/mol)}
    \end{subfigure}
    \hfill
    \begin{subfigure}[b]{0.32\textwidth}
        \centering
        \includegraphics[width=\textwidth]{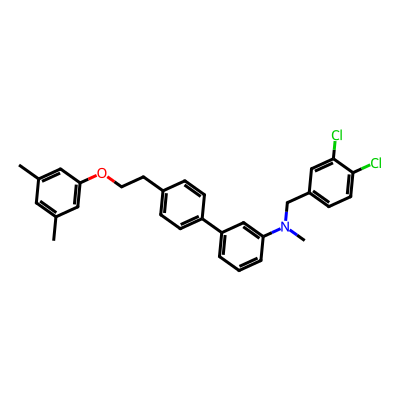}
        \caption{M2 (3UON, -10.9 kcal/mol)}
    \end{subfigure}
    
    \vspace{0.3cm}
    
    \begin{subfigure}[b]{0.32\textwidth}
        \centering
        \includegraphics[width=\textwidth]{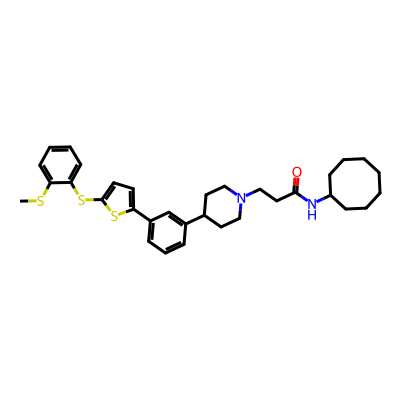}
        \caption{AKT1 (4GV1, -9.4 kcal/mol)}
    \end{subfigure}
    \hfill
    \begin{subfigure}[b]{0.32\textwidth}
        \centering
        \includegraphics[width=\textwidth]{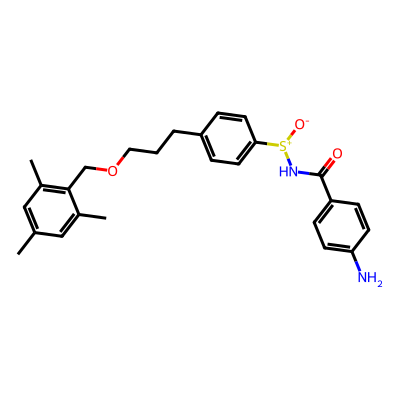}
        \caption{DRD2 (6CM4, -10.7 kcal/mol)}
    \end{subfigure}
    \hfill
    \begin{subfigure}[b]{0.32\textwidth}
        \centering
        \includegraphics[width=\textwidth]{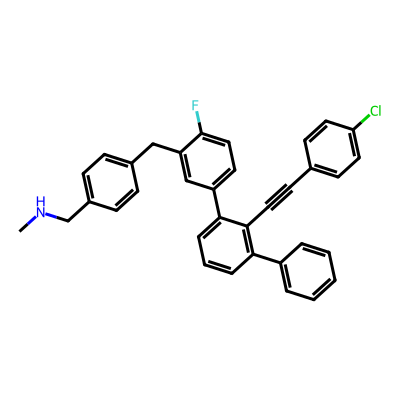}
        \caption{KOR (4DJH, -11.3 kcal/mol)}
    \end{subfigure}
    
    \caption{Representative high-affinity molecules generated by ReACT-Drug for 6 diverse protein targets. All molecules are chemically valid, synthetically accessible, and novel relative to the MOSES benchmark.}
    \label{fig:top_molecules}
\end{figure*}

Table~\ref{tab:sbdd_benchmark} presents a comprehensive comparison of mean binding affinity scores across generalized structure-based drug design models. Despite being entirely target-agnostic, ReACT-Drug achieves competitive mean binding affinities across all six targets, ranging from -9.13 to -10.4 kcal/mol over 5 episodes. The best-performing molecules generated by our framework achieve substantially stronger binding: -11.3 kcal/mol for KOR, -10.9 kcal/mol for M2, and -10.7 kcal/mol for DRD2, as shown in Figure~\ref{fig:top_molecules}.

Our framework outperforms SMILES-based generative models like GVAE~\cite{Kusner2017, Ciep2023}  and CVAE~\cite{Gomez2018, Ciep2023} by substantial margins (approximately 4-5 kcal/mol improvement) on the serotonin and muscarinic receptor targets. Against 3D structure-based models on AKT1, our mean score (-9.13 kcal/mol) surpasses DrugGEN (-8.386 kcal/mol), RELATION (-8.100 kcal/mol), and TargetDiff (-7.981 kcal/mol). Although REINVENT~\cite{Olivecrona2017, Ciep2023} shows better docking results for 5-HT1B, ReACT-Drug performs better for all the other targets. These results are particularly significant given that our framework operates without any target-specific supervised training.

\textbf{Database Baselines.} These represent reference values from existing molecular databases such as ZINC (commercially available compounds) and training set molecules, providing upper-bound estimates achievable through virtual screening of known chemical space. Our best molecules approach or exceed these baselines in several cases (for example, M2, AKT1 and DRD2) demonstrating effective exploration of novel chemical space.

\textbf{CrossDocked2020 Benchmark.} CrossDocked2020~\cite{francoeur2020crossdocked} is a standardized benchmark containing 22.5 million protein-ligand poses created by docking ligands into similar binding pockets across the Protein Data Bank. The aggregate benchmarks represent average or median performance across 100 diverse pockets. State-of-the-art methods include RxnFlow (-8.85 avg) and TacoGFN (-8.82 median). While direct comparison is complicated by differences in evaluation protocols (per-target vs. aggregate statistics), our performance demonstrates competitive binding optimization.

\textbf{KOR Comparison.} For the kappa opioid receptor (KOR), limited published docking-based binding affinity data from comparable generative models precludes direct comparison. Recent work such as ScaRL-P~\cite{zhu2025scaffold} primarily reports predicted bioactivity scores (pIC$_{50}$) derived from QSAR models, which represent a different measurement paradigm than our physics-based AutoDock Vina scores. Direct numerical comparison between these metrics is not methodologically appropriate. Nevertheless, our best KOR candidate achieves -11.3 kcal/mol, demonstrating that ReACT-Drug effectively generates pharmacologically relevant KOR-targeting molecules despite the absence of target-specific training.

\begin{table*}
\centering
\caption{Comparison of mean binding affinity scores (kcal/mol) across generalized structure-based drug design models. ``--'' indicates no published data available for that target. Scoring methods: $^{V}$AutoDock Vina, $^{S}$SMINA/Vinardo, $^{G}$Glide XP.}
\label{tab:sbdd_benchmark}
\resizebox{\textwidth}{!}{
\begin{tabular}{@{}lccccccl@{}}
\toprule
\textbf{Model} & \textbf{5-HT1B} & \textbf{5-HT2B} & \textbf{M2} & \textbf{AKT1} & \textbf{DRD2} & \textbf{KOR} & \textbf{Citation} \\
 & (4IAQ) & (4NC3) & (3UON) & (4GV1) & (6CM4) & (4DJH) & \\
\midrule
\textbf{ReACT-Drug (Results for 5-episodes)} & -9.3 & -9.23 & \textbf{-10.3} & \textbf{-9.13} & \textbf{-9.9} & -10.4 & -- \\
\midrule
\multicolumn{8}{l}{\textit{SMILES-based Generative Models$^{S}$}} \\
\midrule
REINVENT & 9.774 & -8.657 & -9.775 & -- & -- & -- & \cite{Olivecrona2017, Ciep2023} \\
GVAE & -4.955 & -4.641 & -5.422 & -- & -- & -- & \cite{Kusner2017, Ciep2023} \\
CVAE & -4.647 & -4.188 & -4.836 & -- & -- & -- & \cite{Gomez2018, Ciep2023} \\
DeepTarget$^{G}$ & -- & -- & -- & -- & -8.890 & -- & \cite{chen2023deeptarget} \\
\midrule
\multicolumn{8}{l}{\textit{3D Structure-based Generative Models$^{V}$}} \\
\midrule
DrugGEN & -- & -- & -- & -8.386 & -- & -- & \cite{Unlu2025} \\
RELATION & -- & -- & -- & -8.100 & -- & -- & \cite{zhang2023relation} \\
TargetDiff & -- & -- & -- & -7.981 & -- & -- & \cite{guan2023targetdiff} \\
Pocket2Mol & -- & -- & -- & -7.957 & -- & -- & \cite{peng2022pocket2mol} \\
ResGen & -- & -- & -- & -6.468 & -- & -- & \cite{zhang2023resgen} \\
\midrule
% \multicolumn{8}{l}{\textit{Reference Compounds}} \\
% \midrule
% Ergotamine$^{S}$ & -11.22 & -- & -- & -- & -- & -- & \cite{cieplinski2023generative} \\
% Capivasertib$^{V}$ & -- & -- & -- & -9.517 & -- & -- & \cite{Unlu2025} \\
% Ipatasertib$^{V}$ & -- & -- & -- & -9.681 & -- & -- & \cite{Unlu2025} \\
% Risperidone$^{V}$ & -- & -- & -- & -- & -9.4 & -- & \cite{guo2024augmented} \\
% JDTic$^{G}$ & -- & -- & -- & -- & -- & -12.1 & \cite{salasestrada2023kor} \\
% \midrule
\multicolumn{8}{l}{\textit{Database Baselines$^{S}$}} \\
\midrule
ZINC top 1\% & -10.496 & \textbf{-9.833} & -8.802 & -- & -- & -- & \cite{Ciep2023} \\
Training set top 1\% & \textbf{-11.493} & -- & -10.003 & -- & -- & -- & \cite{Ciep2023} \\
Real AKT1 inhibitors$^{V}$ & -- & -- & -- & -8.387 & -- & -- & \cite{Unlu2025} \\
Real DRD2 inhibitors$^{G}$ & -- & -- & -- & -- & -7.753 & -- & \cite{chen2023deeptarget} \\
\midrule
\multicolumn{8}{l}{\textit{CrossDocked2020 Aggregate Benchmarks$^{V}$ (for reference)}} \\
\midrule
RxnFlow & \multicolumn{6}{c}{-8.85 avg (100 pockets)} & \cite{seo2025rxnflow} \\
TacoGFN & \multicolumn{6}{c}{-8.82 med (100 pockets)} & \cite{seo2024tacogfn} \\
DecompDiff & \multicolumn{6}{c}{-8.39 avg (100 pockets)} & \cite{guan2023decompdiff} \\
TargetDiff & \multicolumn{6}{c}{-7.80 avg (100 pockets)} & \cite{guan2023targetdiff} \\
Pocket2Mol & \multicolumn{6}{c}{-7.15 avg (100 pockets)} & \cite{peng2022pocket2mol} \\
\bottomrule
\end{tabular}
}
\end{table*}

\subsection{Top-Performing Generated Molecules}

Figure~\ref{fig:top_molecules} showcases representative high-affinity molecules generated by ReACT-Drug for each of the six protein targets while the complete set of such molecules are shown in Supplementary Figure S1. These molecules demonstrate the framework's ability to generate chemically diverse structures tailored to different binding pocket architectures. Notably, the kappa opioid receptor (KOR) candidate achieved the strongest predicted binding affinity of -11.3 kcal/mol.  The dopamine D2 receptor (DRD2) candidate (-10.7 kcal/mol) significantly outperforms known DRD2 inhibitors (mean -7.753 kcal/mol), while the muscarinic M2 receptor (3UON) candidate (-10.9 kcal/mol) demonstrates strong binding.

\section{Conclusion}
In this work, we propose ReACT-Drug which is a target-agnostic, reaction-template guided RL based framework for novel drug design. In this regard, we have leveraged the capabilities of ESM-2 and ChemBERTa for the encoding of proteins and drugs respectively along with the design of a PPO agent. The various results as reported in the work show that our framework is quite competitive for the generation of novel drug candidates. This is further reflected in the form of mean binding affinities across the six protein targets like 5-HT1B, 5-HT2B , M2, AKT1, DRD2 and KOR, the values ranging from 9.13 to -10.4 kcal/mol. We have also compared our framework with the state-of-the-art as well as baselines and the results are quite encouraging.

Although ReACT-Drug shows competitive results, the current framework is limited to execution over only five episodes due to resource constraints. Nonetheless, the encouraging outcomes suggest that extending the training to more episodes in future work is likely to result in further performance improvements.

\section*{References}
\bibliographystyle{IEEEtran}
\bibliography{Ref}

\end{document}